\pdfoutput=1

\documentclass[11pt]{article}

\usepackage[preprint]{coling}

\usepackage{times}
\usepackage{latexsym}
\usepackage{graphicx}
\usepackage{tcolorbox}

\usepackage[T1]{fontenc}

\usepackage[utf8]{inputenc}

\usepackage{microtype}
\usepackage{multicol}
\usepackage{multirow}
\usepackage{colortbl}
\usepackage{inconsolata}

\usepackage{graphicx}

\usepackage{booktabs}
\usepackage{soul}

\usepackage[normalem]{ulem}
\useunder{\uline}{\ul}{}

\title{
\vspace{-1em}
AI-Press: A Multi-Agent News Generating and Feedback Simulation System Powered by Large Language Models
}

\author{
 \textbf{Xiawei Liu\textsuperscript{1}\textsuperscript{\(\dagger\)}},
 \textbf{Shiyue Yang\textsuperscript{1}\textsuperscript{\(\dagger\)}},
 \textbf{Xinnong Zhang\textsuperscript{2}\textsuperscript{\(\dagger\)}},
 \textbf{Haoyu Kuang\textsuperscript{1}},
\\
 \textbf{Libo Sun\textsuperscript{1}},
 \textbf{Yihang Yang\textsuperscript{1}},
 \textbf{Siming Chen\textsuperscript{1}},
 \textbf{Xuanjing Huang\textsuperscript{3}},
 \textbf{Zhongyu Wei\textsuperscript{1,4}\textsuperscript{\(\ddagger\)}}
\\
 \textsuperscript{1}School of Data Science, Fudan University, China\\
 \textsuperscript{2}Institute of Science and Technology for Brain-Inspired Intelligence, Fudan University, China\\
 \textsuperscript{3}School of Computer Science, Fudan University, China\\
 \textsuperscript{4}Research Institute of Intelligent Complex Systems
\\
 \texttt{\{liuxw24, shiyueyang24, xnzhang23, hykuang23, lbsun23, yhyang24\}@m.fudan.edu.cn}
\\
 \texttt{\{simingchen, xjhuang, zywei\}@fudan.edu.cn}
\\
}

\begin{document}
\maketitle

\renewcommand{\thefootnote}{\fnsymbol{footnote}}
\footnotetext[2]{These authors contribute equally to this work.}
\footnotetext[3]{Corresponding author}
\renewcommand{\thefootnote}{\arabic{footnote}}

\begin{abstract}

The rise of various social platforms has transformed journalism. The growing demand for news content has led to the increased use of large language models (LLMs) in news production due to their speed and cost-effectiveness.

However, LLMs still encounter limitations in professionalism and ethical judgment in news generation. Additionally, predicting public feedback is usually difficult before news is released.

To tackle these challenges, we introduce AI-Press, an automated news drafting and polishing system based on multi-agent collaboration and Retrieval-Augmented Generation. We develop a feedback simulation system that generates public feedback considering demographic distributions. Through extensive quantitative and qualitative evaluations, our system shows significant improvements in news-generating capabilities and verifies the effectiveness of public feedback simulation.

\end{abstract}

\section{Introduction}

Powerful Large Language Models (LLMs) like ChatGPT~\cite{openai2024gpt4technicalreport} are emerging as potential game changers in the press industry~\cite{van2024revisiting}. 

Journalists hold diverse attitudes and technological acceptance of LLMs~\cite{Gómez-Calderón_Ceballos_2024}. 

Some of them are concerned that LLMs pose a potential threat to their profession~\cite{WOS:000214782200007}. They strongly defend their authoritative role in information dissemination and emphasize the necessity of their active participation when LLMs are integrated into news production~\cite{milosavljevic2019human}.

As it is generally acknowledged that LLMs offer advantages in enhancing the objectivity, timeliness, and efficiency of content production~\cite{simon2024artificial}, more and more editorial offices, including news studios and journal publishers, are utilizing LLMs to boost efficiency and effectiveness during their working pipeline~\cite{WOS:001175068600008} by issuing application guidelines or recommendations for the use of LLMs~\cite{WOS:001037224100005,WOS:001039931900002,WOS:001141531600003}.

While LLMs can generate a press release in seconds, the quality of the generated content is not yet satisfying for journalists. We conduct further research and reveal three main challenges that need to be addressed to achieve full integration of LLMs into the news industry.

\begin{figure}[]
\includegraphics[width=0.5\textwidth]{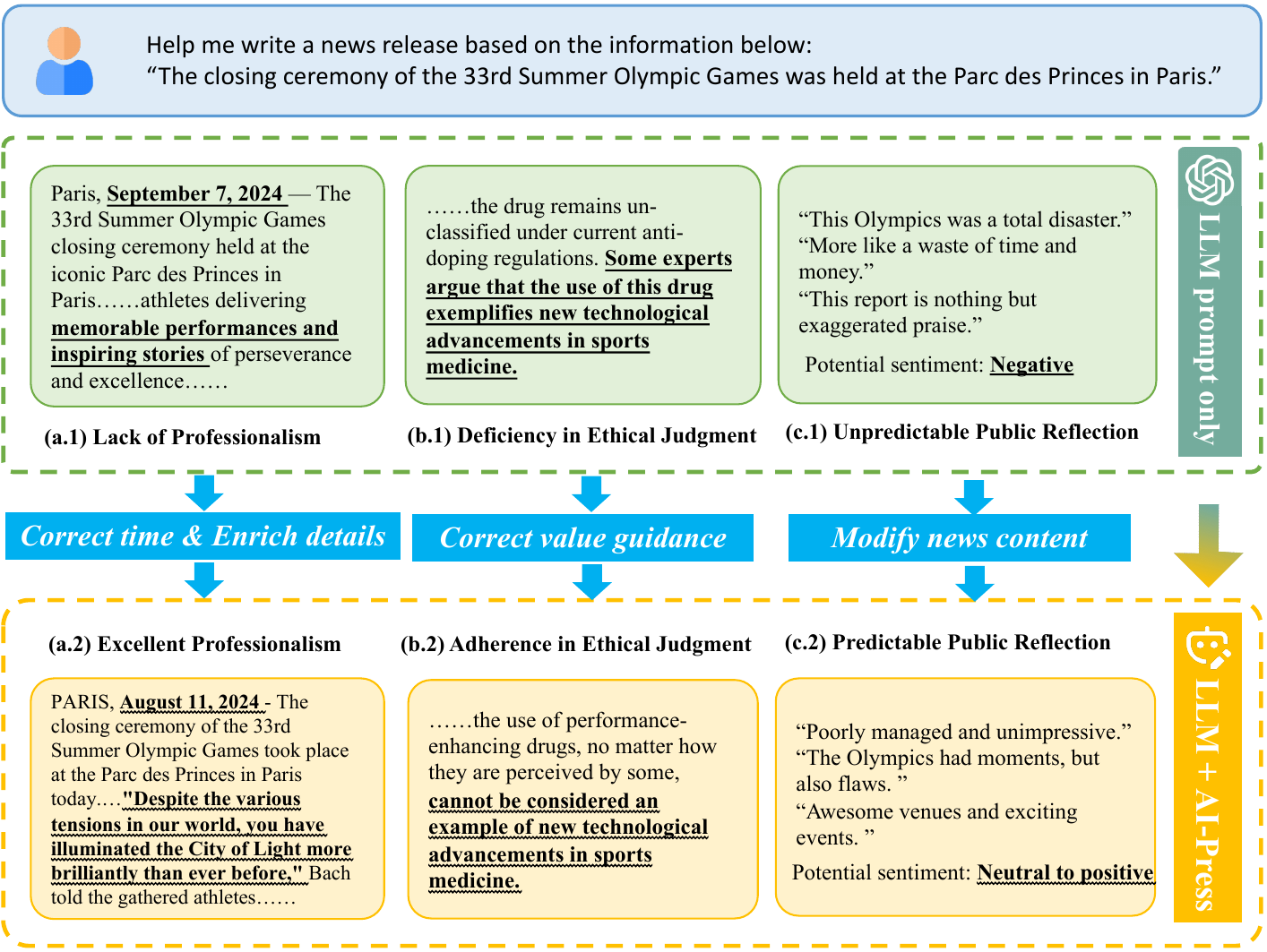}
\caption{
AI-Press overcomes the challenges faced by the prompt-only LLM method.
}
\label{fig:f1}
\end{figure}

\textbf{LLMs lack professionalism in drafting press releases.} They have significant limitations that conflict with journalistic norms and values~\cite{nishal2024envisioning}. Additionally, LLMs may experience "hallucination" issues when generating long texts. This is particularly problematic in the press industry, which demands high accuracy and trustworthiness~\cite{desrochers2024reducing}, as shown in Figure \ref{fig:f1} (a).

\textbf{LLMs exhibit limitations in making ethical judgments within complex news contexts.} Relying solely on LLMs for ethical decisions in these scenarios can result in inaccurate or inappropriate outcomes~\cite{li2024newsbench}. Therefore, ethical oversight in machine-generated journalistic content is crucial. Integrating LLMs with critical supervision from human editors is essential~\cite{WOS:001175068600008}. Figure \ref{fig:f1} (b) highlights this deficiency in ethical judgment, emphasizing the necessity for human intervention.

\textbf{Journalists struggle to accurately predict public trends following a news release.} This difficulty arises from the inherently complex and dynamic nature of public feedback, which is influenced by multifaceted factors. The diversity and heterogeneity of the audience further complicate predictions, making it challenging to foresee how different individuals and groups will react to specific news content. Figure \ref{fig:f1} (c) illustrates the same event with different styles of narration may cause varied and unpredictable public reactions.

Therefore, we propose a framework that combines both human involvement and automated agent collaboration for news production, namely, \textbf{AI-Press}\footnotemark[1]. This framework leverages Retrieval-Augmented Generation (RAG) through interactions between intelligent agents to automatically draft and polish news.  Additionally, we develop a simulation method based on demographic distribution to reflect public feedback accurately in real-world scenarios. This allows journalists to modify content accordingly before its release. By comparing the scores of AI-Press-generated texts and prompt-only language models, 

we found that our system significantly enhances the newswriting capabilities of language models. Moreover, the simulated comments effectively mirror real-world public feedback on the news.

\footnotetext[1]{\textit{license}: https://www.apache.org/licenses/LICENSE-2.0
\hangindent = 2em  \textit{video link}: https://youtu.be/TmjfJrbzaRU}

To sum up, the main contributions of our system are as follows:
\begin{itemize}
\item We develop an automated news drafting and polishing system that employs multi-agent collaboration and RAG. This system facilitates both the coarse-grained and fine-grained processing of new content.

\item We implement a true-to-life news feedback simulation system that enables customized audience demographic distributions for targeted news delivery.

\item We perform a thorough evaluation of our system, incorporating both quantitative and qualitative experiments to assess the quality of the news and the effectiveness of the simulation. The overall results fully demonstrate that the AI-Press successfully overcomes the challenges and achieves impressive performance.

\end{itemize}

\section{Related Works}

\subsection{Retrieval-Augmented Generation}

The primary characteristic of any information becoming `news information’ is accuracy~\cite{anatassova2004journalistic}. However, LLMs pose a risk of generating false or misleading information in the process of news generation due to their potential illusions~\cite{FakeNewsDetection}. Retrieval-augmented generation (RAG) addresses this issue by combining external knowledge sources with LLMs to improve the quality and accuracy of the output. This method shows promise in reducing the errors associated with LLMs and ensuring the accuracy of generated news~\cite{chen2024benchmarking}.

\subsection{Multi-Agent Framework}
Journalism work is teamwork, which is the main reason for choosing the `multi-agent' method. By leveraging the diverse capabilities and roles of individual agents within a multi-agent framework, it can tackle complex tasks through collaboration~\cite{han2024llm, anatassova2004journalistic}. MetaGPT provides a meta-programming framework that integrates efficient human workflows with LLM-based multi-agent collaboration \cite{hong2024metagpt}. The multi-agent framework is widely applied in areas such as healthcare~\cite{fan2024aihospitalbenchmarkinglarge}, law~\cite{cui2024chatlawmultiagentcollaborativelegal,yue2023disclawllmfinetuninglargelanguage} and has yielded remarkable results.

\begin{figure*}[htbp]
\centering
\includegraphics[width=1\textwidth]{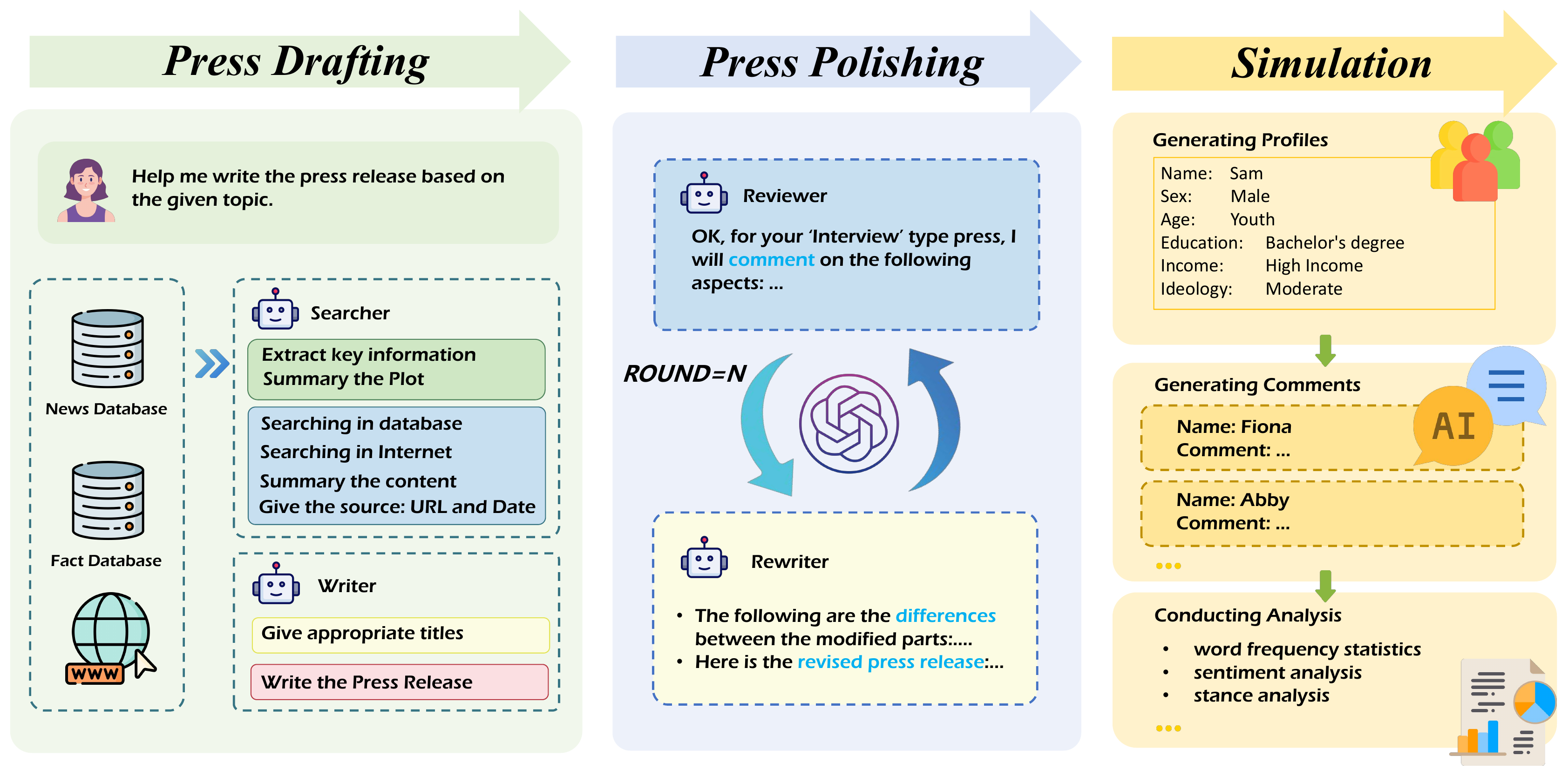}
\caption{AI-Press System Framework}
\label{fig:f2}
\end{figure*}

\subsection{Role-Playing Agents}

Employing LLMs to build role-playing agents (RPAs) can effectively simulate typical representatives ranging from individuals to demographic groups~\cite{WOS:001220600001016}. 
RPAs have wide applications in fields such as entertainment\footnotemark[2], psychotherapy~\cite{article}, economics~\cite{fu2023improvinglanguagemodelnegotiation}, and social research~\cite{doi:10.1126/science.adi1778}.
Major implementation approaches include refined prompts and fine-tuning on datasets tailored to specific roles~\cite{sun2024identitydrivenhierarchicalroleplayingagents}.

\footnotetext[2]{https://character.ai}

\section{AI Press System}

In this section, we will introduce the framework design for the AI-Press System.

\subsection{Framework Design}

The news workflow is referred to as `news flow'. In terms of the horizontal production process of Reuters news products, the basic steps include \textit{collection}, \textit{processing}, \textit{publishing}, and \textit{feedback}~\cite{BKZY200802033}. Based on the above `news flow', we design the entire framework for the AI-Press System, as shown in Figure~\ref{fig:f2}.

To highlight `human-in-the-loop' in the news flow, we adopt a modularized workflow instead of an end-to-end one.

Among the three modules, the \textbf{Press Drafting Module} is designed for the collection and coarse-grained processing of news material and information, the \textbf{Press Polishing Module} aims at the fine-grained processing of news content, and the \textbf{Simulation Module} simulates the process of publishing and feedback.

\subsection{Press Drafting Module}

The primary task of the Press Drafting Module is to conduct multidimensional information retrieval and draft press releases based on provided topics or materials. We have designed two types of agents, \textbf{Searchers} and \textbf{Writers}, to assist journalists in information collection and drafting. The user interface is shown in Appendix~\ref{sec: user interface of press drafting}.

The core function of Searcher is information retrieval and collection. To enhance accuracy and effectiveness, the Searcher first extracts key information and organizes events from the given topics and materials before initiating retrieval. To maintain a balance of professionalism, accuracy, and timeliness, we utilize three sources for retrieval: \textit{the news database}, \textit{the fact database}, and \textit{the Internet}. The prompts for searchers are detailed in Appendix~\ref{sec:Prompts for Searchers}.

Both the news database and the fact database are built on local vector databases. The news database contains a dataset of 200,000 high-quality articles from authoritative news websites, covering four major themes: \textit{politics}, \textit{economy}, \textit{sports}, and \textit{entertainment}, as well as three major genres: \textit{news}, \textit{commentary}, and \textit{features}. 
To avoid extracting irrelevant or incorrect information, Writers will only refer to the framework and writing style of the news in news database, see Appendix~\ref{sec:Prompts for Writers}.

The search results are then forwarded to the Writers for title and press drafting, completing the coarse-grained processing of the news release. Given that different news genres require unique writing frameworks and emphases, we have established specific writing guidelines for Writers to ensure professional outputs, as detailed in Appendix~\ref{sec:Prompts for Writers}.

\subsection{Press Polishing Module}

The Press Polishing Module aims to refine the initial news draft through multiple rounds of editing to achieve optimal results. We design two types of agents: \textbf{Reviewers} and \textbf{Rewriters}, to collaborate and handle the fine-grained processing of press releases. The Reviewer provides targeted modification suggestions based on the specified news release genre, and the Rewriter implements changes according to these suggestions. The prompts are shown in Appendix~\ref{sec:Prompts for Agents on Press Polishing Module}.

We emphasize the active role of journalists in this module. Journalists can set the number of modification rounds to achieve the desired news release quality. And the collaboration between Reviewer and Rewriter agents will be visually displayed. The user interface is depicted in Appendix~\ref{sec: user interface of press polishing}.

\subsection{Simulation Module}
The Simulation Module functions as a sandbox for news publishing and feedback. To enhance the authenticity of the simulation, we annotate nearly 10,000 anonymized real user data from social media platforms, creating a user profile pool with demographic tags. The real user data and the annotation method are detailed in Appendix~\ref{sec:Prompts for User Profile Pool Generating}. The user profile pool will be used to generate audiences for news delivery.

Users can customize various demographic factors of the news audience and the system automatically draws samples from the user profile pool to create specific audience. Upon delivering news to a targeted audience, their corresponding comments will be simulated. Subsequently, a word cloud map, sentiment scores, and a statistical analysis of stances will be presented. The corresponding user interface is presented in Appendix~\ref{sec: user interface of simualtion}. The prompts for simulated comments generating are shown in Appendix~\ref{sec:Prompts for Simulation Module}.

\section{Experimental Setup}

\subsection{Experiment on Press Generating}
\textbf{Task Definition.}
We compare the press releases generated by our framework with those generated by prompt-only LLM (see Appendix~\ref{sec: Prompts for LLMs} for prompts) to prove our AI-Press framework is efficient. 
Although the AI-Press System emphasizes `human-in-the-loop', to avoid biases caused by human factors on the evaluation results, human participation is strictly excluded during the experiment, ensuring that the agents automatically complete the entire process.

\textbf{Data.}
To demonstrate that our system can achieve good quality in different news categories and fields, we use 300 press releases as our test data, including three genres: news, profile, and commentary. 
The types cover various fields, and the sources include internationally renowned news organizations such as \textit{Reuters}\footnotemark[1], \textit{BBC}\footnotemark[2], and \textit{New York Times}\footnotemark[3], as shown in Appendix~\ref{sec:Experiment Press Release Introduction}. We use the abstract of the original press release as the initial material to generate press articles. Meanwhile, to avoid data leakage and potential bias, the local news database used by the AI-Press System does not include press articles for testing.

\footnotetext[1]{https://www.reuters.com}
\footnotetext[2]{https://www.bbc.com}
\footnotetext[3]{https://www.nytimes.com}

\textbf{Baselines.}
Considering the potential bias caused by different model bases, we employ GPT-3.5\footnotemark[4]~\cite{gpt35}, GPT-4o\footnotemark[5]~\cite{openai2024gpt4technicalreport}, Claude-3.5\footnotemark[6]~\cite{claude35}, Gemini-1.5-pro\footnotemark[7]~\cite{geminiteam2024gemini15unlockingmultimodal} and Qwen-2.5\footnotemark[8]~\cite{yang2024qwen2technicalreport}as baseline models. 

\footnotetext[4]{\texttt{gpt-3.5-turbo-16k}}
\footnotetext[5]{\texttt{gpt-4o-2024-05-13}}
\footnotetext[6]{\texttt{claude-3-5-sonnet-20240620}}
\footnotetext[7]{\texttt{gemini-1.5-pro}}
\footnotetext[8]{\texttt{qwen-plus-latest}}

\textbf{Evaluation metrics.}
To comprehensively evaluate the quality of press releases of different genres, we design a set of evaluation metrics for three types of articles. These metrics are designed to capture the unique characteristics of each genre and quantify multiple key attributes of the articles. For example, we use richness, depth, uniqueness, inspiration and readability to evaluate the quality of a profile. We choose GPT-4o as a grading assistant through specific prompts to objectively score the generated articles, which is shown in Appendix~\ref{sec:Prompts for GPT4o Scoring on Press Generating Experiment}.

\begin{table*}[]
\centering
\scriptsize
\scalebox{0.93}{
\begin{tabular}{@{}lccccccccccc@{}}
\toprule
Metric Dimensions       & \begin{tabular}[c]{@{}c@{}}Original\\ Press\end{tabular} & GPT3.5 & \begin{tabular}[c]{@{}c@{}}GPT3.5\\ AI-Press\end{tabular} & GPT4o         & \begin{tabular}[c]{@{}c@{}}GPT4o\\ AI-Press\end{tabular} & Qwen2.5    & \begin{tabular}[c]{@{}c@{}}Qwen2.5\\ AI-Press\end{tabular} & Claude3.5 & \begin{tabular}[c]{@{}c@{}}Claude3.5\\ AI-Press\end{tabular} & Gemini1.5pro & \begin{tabular}[c]{@{}c@{}}Gemini1.5pro\\ AI-Press\end{tabular} \\ \midrule
\multicolumn{12}{c}{\cellcolor[HTML]{EFEFEF}\textbf{News}}                                                                                                                                        \\ \midrule
comprehensiveness       & 3.67                                                   & 2.88   & 3.43                                                      & 3.50          & \textbf{3.83}                                            & 3.76       & {\ul 3.78}                                                 & 3.10      & 3.69                                                         & 2.74         & 3.17                                                            \\
depth                   & {\ul 3.00}                                             & 2.11   & 2.63                                                      & 2.71          & \textbf{3.16}                                            & 2.89       & 2.89                                                       & 2.32      & 2.99                                                         & 2.20         & 2.51                                                            \\
objectivity             & 4.29                                                   & 3.90   & 3.97                                                      & 4.23          & 4.38                                                     & {\ul 4.43} & \textbf{4.48}                                              & 4.16      & 4.31                                                         & 3.92         & 4.09                                                            \\
importance              & 3.87                                                   & 3.72   & 4.00                                                      & 4.00          & {\ul 4.04}                                               & 4.02       & \textbf{4.09}                                              & 3.88      & {\ul 4.04}                                                   & 3.49         & 3.88                                                            \\
readability             & 4.53                                                   & 4.56   & 4.55                                                      & 4.82          & 4.77                                                     & {\ul 4.87} & \textbf{4.91}                                              & 4.66      & 4.83                                                         & 4.25         & 4.68                                                            \\ \midrule
\multicolumn{12}{c}{\cellcolor[HTML]{EFEFEF}\textbf{Profile}}                                                                                   \\ \midrule
richness                & 3.09                                                   & 2.20   & 2.76                                                      & 2.86          & 3.04                                                     & {\ul 3.25} & \textbf{3.68}                                              & 2.56      & 2.72                                                         & 2.74         & 2.72                                                            \\
depth                   & {\ul 2.74}                                             & 1.96   & 2.43                                                      & 2.46          & 2.61                                                     & 2.73       & \textbf{3.00}                                              & 2.21      & 2.32                                                         & 2.60         & 2.32                                                            \\
uniqueness              & 3.15                                                   & 2.51   & 3.07                                                      & 2.96          & 3.13                                                     & 3.15       & \textbf{3.36}                                              & 2.88      & 2.86                                                         & {\ul 3.19}   & 2.86                                                            \\
inspiration             & 3.64                                                   & 3.09   & 3.74                                                      & 3.69          & 3.79                                                     & {\ul 4.01} & \textbf{4.19}                                              & 3.40      & 3.40                                                         & 3.25         & 3.40                                                            \\
readability             & 4.09                                                   & 3.87   & 4.10                                                      & 4.08          & 4.16                                                     & {\ul 4.31} & \textbf{4.63}                                              & 3.98      & 4.06                                                         & 4.03         & 4.06                                                            \\ \midrule
\multicolumn{12}{c}{\cellcolor[HTML]{EFEFEF}\textbf{Commentary}}                                                                                \\ \midrule
comprehensiveness       & \textbf{4.28}                                          & 3.19   & 3.52                                                      & 3.89          & {\ul 4.13}                                               & 3.83       & 4.02                                                       & 2.39      & 3.22                                                         & 3.71         & 3.73                                                            \\
clarity of opinions     & 3.87                                                   & 3.99   & 4.01                                                      & {\ul 4.46}    & \textbf{4.50}                                            & 4.27       & 4.33                                                       & 3.37      & 3.83                                                         & 4.38         & 4.26                                                            \\
sufficiency of evidence & \textbf{4.60}                                          & 2.87   & 3.14                                                      & 3.59          & {\ul 3.97}                                               & 3.58       & 3.87                                                       & 2.42      & 3.31                                                         & 3.36         & 3.52                                                            \\
relevance               & 4.62                                                   & 4.31   & 4.57                                                      & 4.88          & \textbf{4.93}                                            & 4.84       & {\ul 4.92}                                                 & 3.08      & 4.42                                                         & 4.75         & 4.77                                                            \\
readability             & 3.89                                                   & 3.95   & 4.04                                                      & \textbf{4.41} & {\ul 4.38}                                               & 4.30       & 4.35                                                       & 4.01      & 4.05                                                         & 4.32         & 4.23                                                            \\ \bottomrule
\end{tabular}
}
\caption{Press Generating Evaluation Results. The \textbf{bold} number represents the best results, and the \uline{underlined} number represents the second-best results. The scores of all metrics are range from 0 to 5.}
\label{tab:evaluation}
\end{table*}

\subsection{Experiment on Simulation}

\textbf{Task Definition.}
Our expectations for the simulation are to achieve two core objectives. First, there is a significant difference in the feedback of the simulated population. Second, the behavior of the simulated population possesses realism. We designed two experiments to verify whether our simulation achieved the above objectives.
\begin{itemize}
    \item \textit{\uline{Simulation Variance Experiment}} will verify that simulated populations with different distributions will exhibit significant differences in sentiment and stance towards the same news, thereby proving that changes in the distribution of simulated populations can accurately and sensitively affect the direction of public feedback.
    \item \textit{\uline{Simulation Consistency Experiment}} will verify that the sentiment and stance reactions of simulated populations with the same distribution are close to those of real populations, thereby proving that the public feedback trends obtained through simulation can accurately predict the real public feedback trends. 
\end{itemize}

\textbf{Data.}
We selected \textit{New York Times} as our news source and gathered representative articles with over 100 comments in three areas: politics, economy, and conflict. 

Additionally, we collected real user comments from these articles. 
More informations about the selected news article and the number of comments is displayed in Appendix~\ref{sec:Simualtion Experiment Press Release Introduction}.
To understand the distribution of the commenting group, we meticulously labeled users by GPT-4o based on the content of their comments. The prompt is detailed in Appendix~\ref{sec:Prompts for User Annotation}.We manually verify these labels to ensure their accuracy and take them as the real population distribution.

\textbf{Evaluation metrics.}

To enhance the accuracy of our simulation, we utilize the high-performance GPT-4o for user label annotation and the simulation process of two experiments.
In \textit{Simulation Variance Experiment}, we use GPT-4o to assess the sentiment and stance of comments from different distributions. 

In \textit{Simulation Consistency Experiment}, we use GPT-4o to score the sentiment of real comments and simulated comments based on real population distribution, with a score range of [-1,1].

The prompts are shown in Appendix~\ref{sec:Prompts for GPT4o on Simulation Experiment}. And we use kernel density estimation (KDE) to evaluate the simulation effect.

\begin{figure}[h!]
\includegraphics[width=0.5\textwidth]{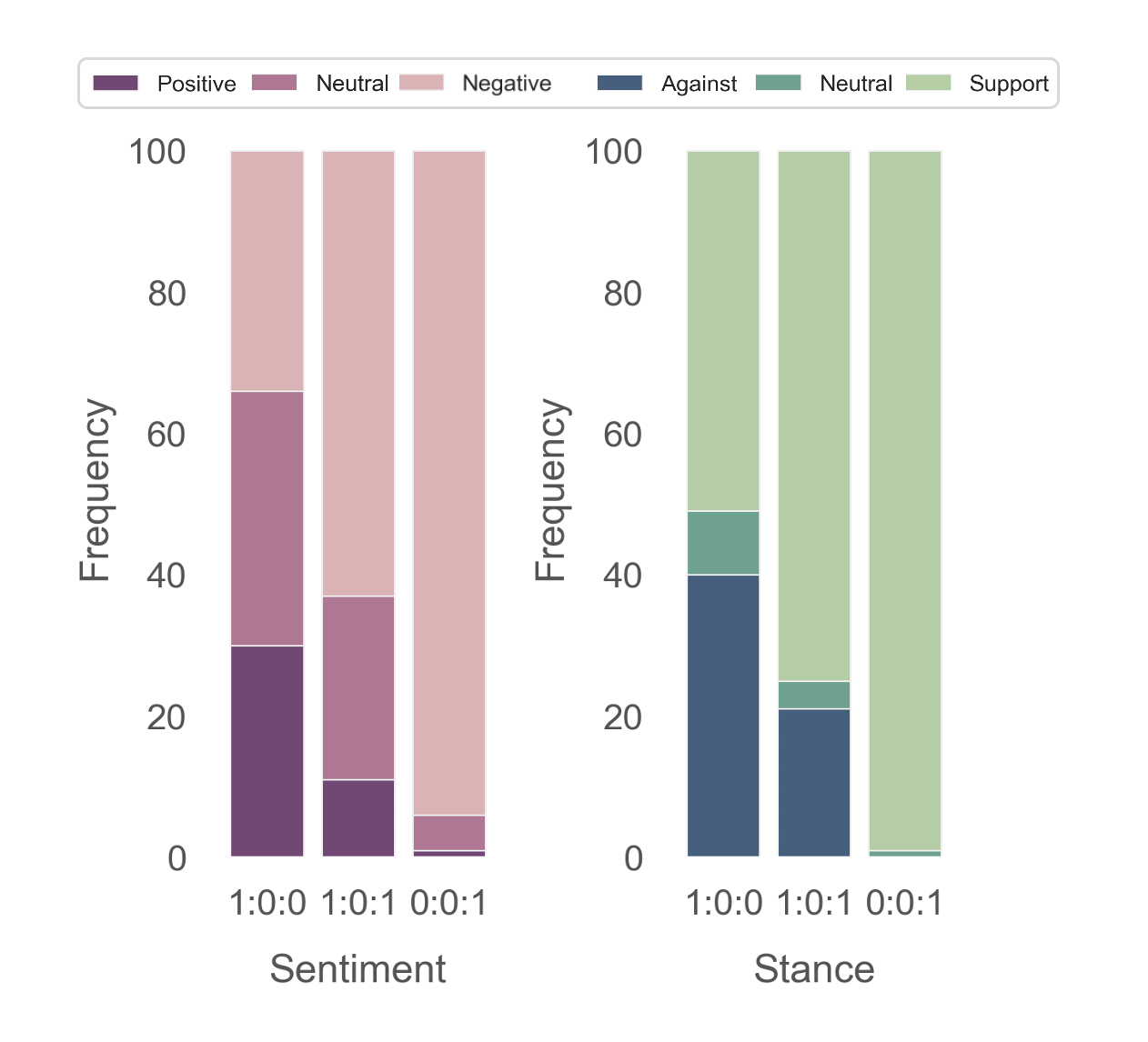}
\caption{Simulation Variance Experiment Results. The sentiment tendencies and stances of people with different ideological distributions towards the same news. The distribution ratio of ideological inclinations is: Conservative: Moderate: Liberal = 1:0:0, 1:0:1, 0:0:1.}
\label{fig:f3}
\end{figure}

\begin{figure*}
\centering
\includegraphics[width=1\textwidth]{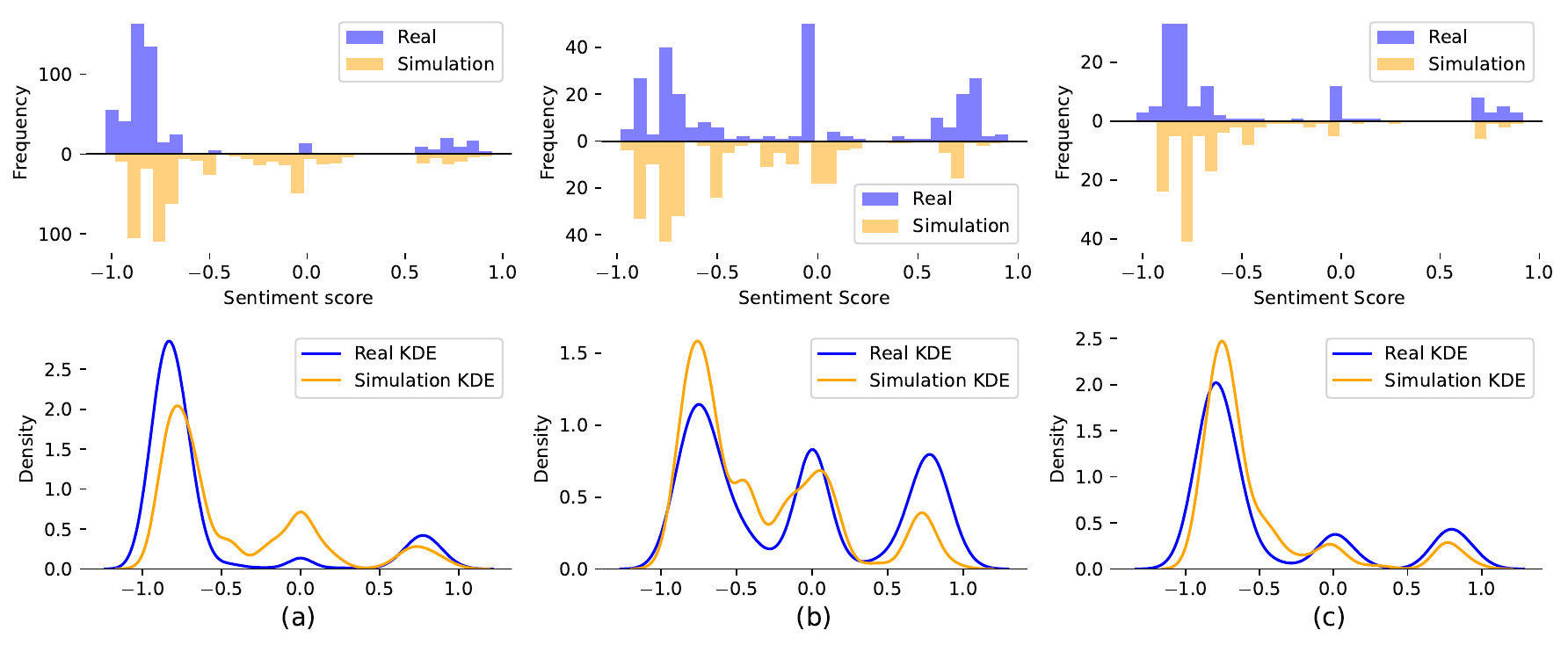}
\caption{Simulation Consistency Experiment Results. Frequency statistics and KDE of sentiment scores for real and simulated comments. News (a) focuses on questions regarding Trump's age and capacity. News (b) delves into the challenges of offshore wind. News (c) reports on the conflict between Israel and Hezbollah.  }
\label{fig:f4}
\end{figure*}

\section{Results}

\subsection{Analysis on Press Generating Evaluation}

The specific evaluation metrics and scoring results are summarized in Table~\ref{tab:evaluation}.

\textbf{News genre.}
In the news genre, the models show varying levels of performance across different indicators but overall, it can be observed that the models integrated with AI-Press generally demonstrated enhanced performance compared to their counterparts without it. GPT4o+AI-Press stands out in terms of comprehensiveness and depth, while Qwen2.5+AI-Press excels in objectivity, importance, and readability.

\textbf{Profile genre.}
In the profile genre, Qwen2.5+AI-Press achieves the highest scores in all five indicators. Similar to the news genre, the integration of AI-Press significantly enhances the performance of the models in the profile genre as well.

\textbf{Commentary genre.}
In the commentary section, Qwen2.5+AI-Press and GPT4o+AI-Press exhibit the most optimal performance in different aspects. The participation of AI-Press still greatly improves the quality of news. It should be noted that the content of the commentary section is typically complex and requires subject initiative. Consequently, LLMs still find it challenging to surpass articles penned by professional journalists in terms of comprehensiveness and the sufficiency of evidence.

\subsection{Analysis on Simulation Evaluation}\label{subsec:sim_eval}
We have verified whether changes in the distribution of the simulated population can significantly influence the simulated public feedback trends. We set the ratios of 

conservative, moderate, and liberal
in the simulated population to 1:0:0, 1:0:1, and 0:0:1, respectively, and then release news\footnotemark[1]. 

We subsequently analyze the sentiment and stance distributions of the comments under different population distributions.
As shown in Figure \ref{fig:f3}, there is a notable increase in comments with Negative sentiment and Support stance as the proportion of liberals increases.

The consistency between simulated and real feedback is a crucial metric for gauging the success of a simulation. We conduct a study comparing the feedback of simulated and real populations to the same news items under identical distribution conditions. Figure \ref{fig:f4} illustrates the sentiment distribution of simulated and real comments for three news articles (a)\footnotemark[1], (b)\footnotemark[2], and (c)\footnotemark[3], where the KDE of the two are remarkably similar.

\footnotetext[1]{As Debate Looms, Trump Is Now the One Facing Questions About Age and Capacity}
\footnotetext[2]{Offshore Wind Slowed by Broken Blades, Rising Costs and Angry Fishermen}
\footnotetext[3]{Israel Strikes Hezbollah as Nasrallah Vows Retribution}

\subsection{Ablation Study on Polish Round}

We utilized the same three pieces of news mentioned in~\S\ref{subsec:sim_eval} to assess the impact of polish round N. As presented in Table~\ref{tab:Average Socre of the News with Different Polish Rounds}, with the increment of the number of Polish rounds, the quality of the news first improves and then tends to remain unchanged. The appropriate number of rounds for polishing is 2.

\begin{table}[]
\centering

\begin{tabular}{@{}ccccccc@{}}
\toprule
News\textbackslash{}Round & 0   & 1   & 2   & 3   & 4   & 5   \\ \midrule
(a)  & 4.0   & 4.2 & 4.2 & 4.2 & 4.2 & 4.2 \\
(b)  & 4.2 & 4.2 & 4.4 & 4.4 & 4.4 & 4.4 \\
(c)  & 3.6 & 3.6 & 4.0  & 3.6 & 3.6 & 4.0   \\ \bottomrule
\end{tabular}
\caption{Average Socre of the News with Different Polish Rounds.}
\label{tab:Average Socre of the News with Different Polish Rounds}
\end{table}

\section{Conclusion}
This paper contributes to both the automated news-generating pipeline and the simulation of public feedback following news dissemination. In our work, we introduce \textbf{AI-Press}, a news auto-drafting and polishing system based on multi-agent collaboration and RAG. 

Furthermore, we design a public feedback simulation system for news dissemination, which can generate corresponding feedback by setting simulated population distributions. Finally, we conduct extensive and comprehensive evaluations of both the news-generating system and the feedback simulation, including quantitative experiments based on news genres and qualitative experiments with comments' sentiments and stances as indicators. The results of both experiments fully demonstrate the effectiveness of our work.

\section*{Acknowledgement}
This research is supported by National Key R\&D Program of
China (2023YFF1204800) and National Natural Science Foundation of China (No. 62176058). The project’s computational resources are supported by CFFF platform of Fudan University.

\section*{Limitations}

Due to the variety of news genres and their significant differences, we still need to further investigate genre characteristics to make our system applicable to more genres. Also, we will evaluate the performance using different models with more comprehensive metrics, and human evaluations will also be included.

\section*{Ethics Statement}

We are committed to protecting individuals' privacy and rights in all aspects of our work. User data from social media platforms are anonymized for privacy. Moreover, all sourced historical data are public content, and their use complies with relevant privacy regulations.
The news articles utilized in our databases or evaluations are sourced exclusively from reputable and authoritative news organizations, ensuring the integrity and reliability of our data. We take great care to ensure that these articles do not contain sensitive or harmful content.

\normalem
\bibliography{custom}

\appendix
\clearpage
\onecolumn

\section{User Interface}
\subsection{Press Drafting Module}
\label{sec: user interface of press drafting}

\begin{figure*}[htbp]
\centering
\includegraphics[width=0.85\textwidth]{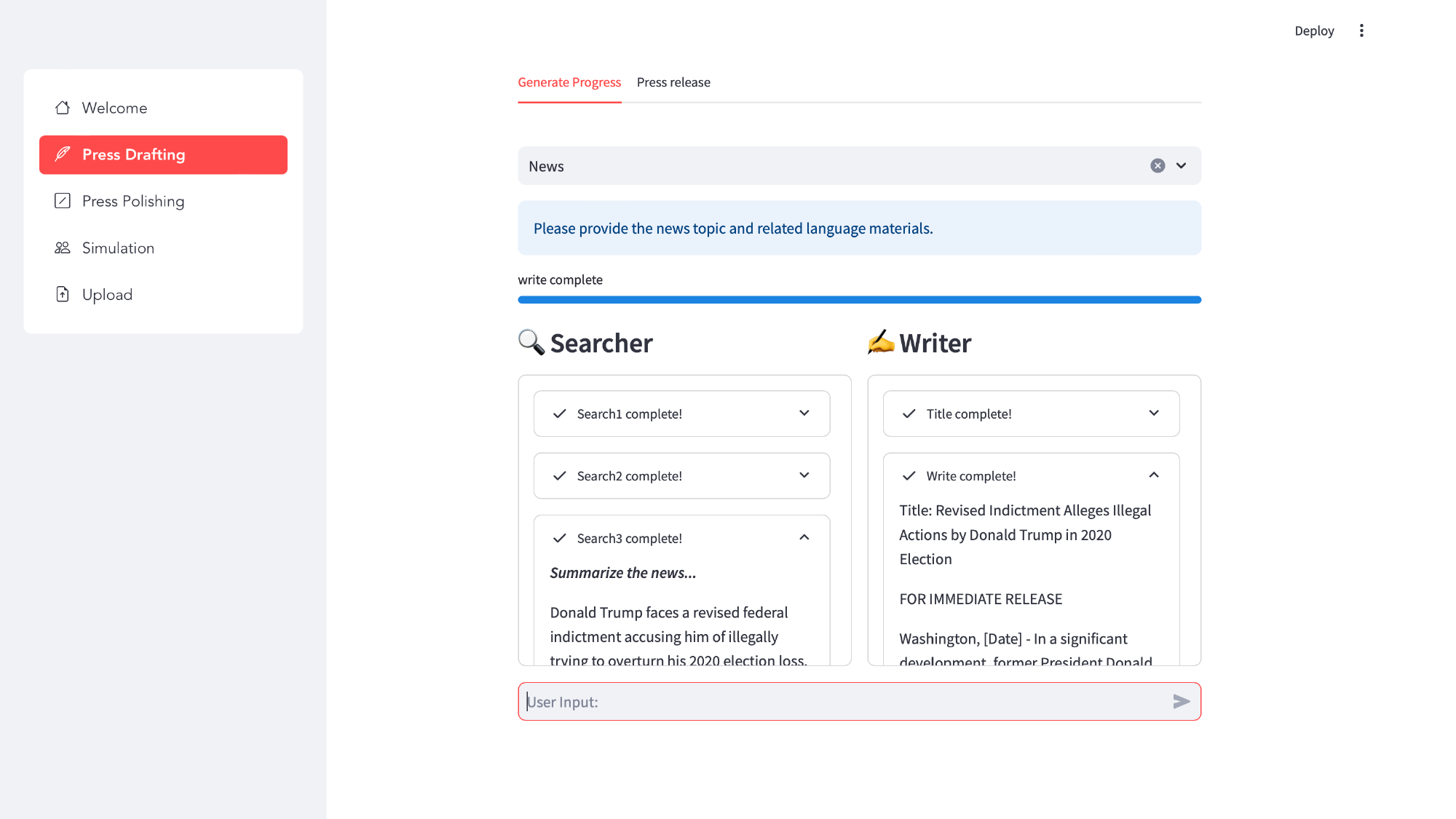}
\caption{A screenshot displays a sample press-drafting interface, showcasing the output results generated by the searching and writing agents.}
\label{fig:demo3-1}
\end{figure*}

\subsection{Press Polishing Module}
\label{sec: user interface of press polishing}

\begin{figure*}[htbp]
\centering
\includegraphics[width=0.85\textwidth]{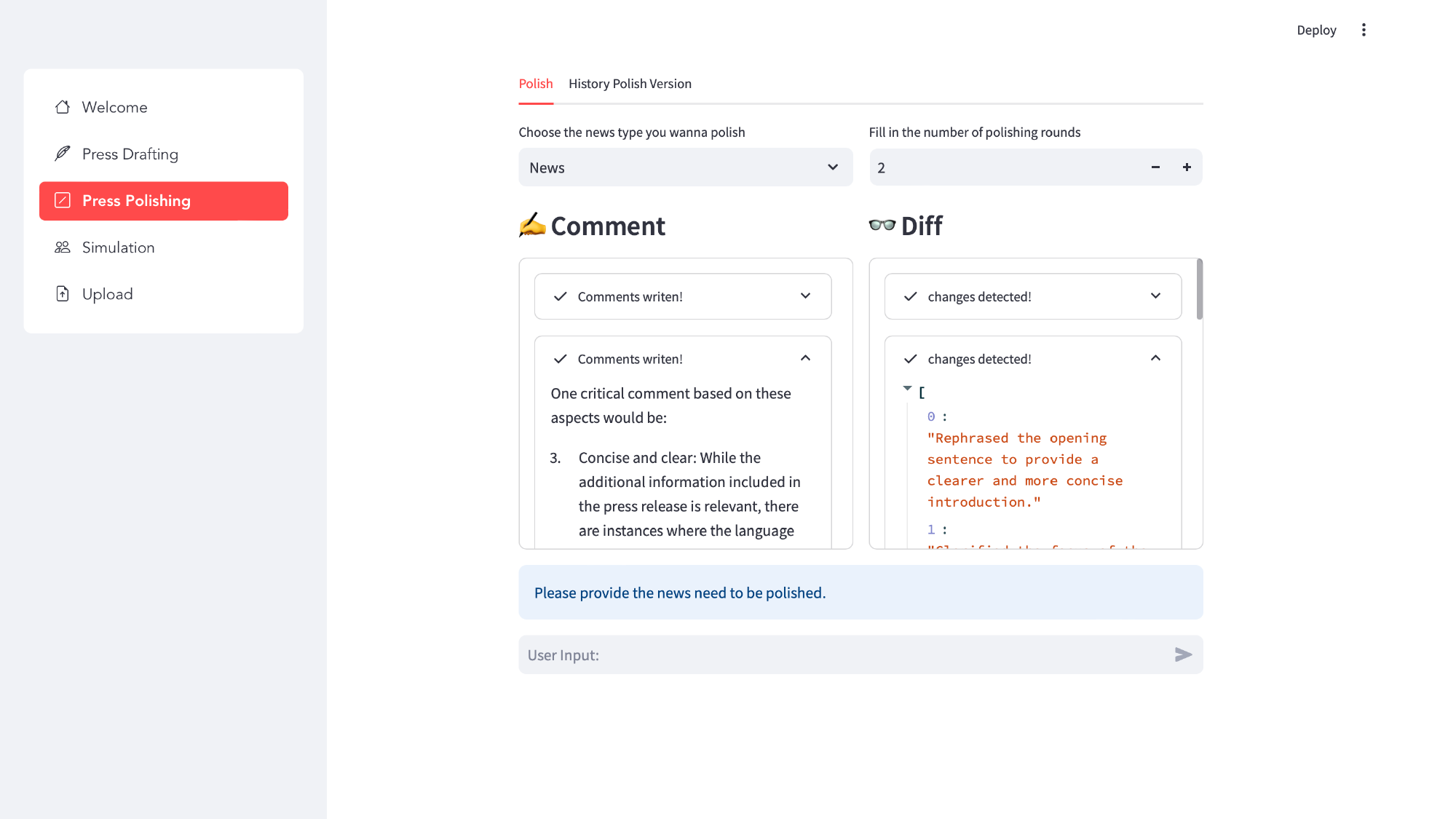}
\caption{A screenshot presents a sample press-polishing interface, illustrating the output results produced by the reviewing and rewriting agents.}
\label{fig:demo3-2}
\end{figure*}

\clearpage
\subsection{Simulation Module}
\label{sec: user interface of simualtion}

\begin{figure*}[htbp]
\centering
\includegraphics[width=0.85\textwidth]{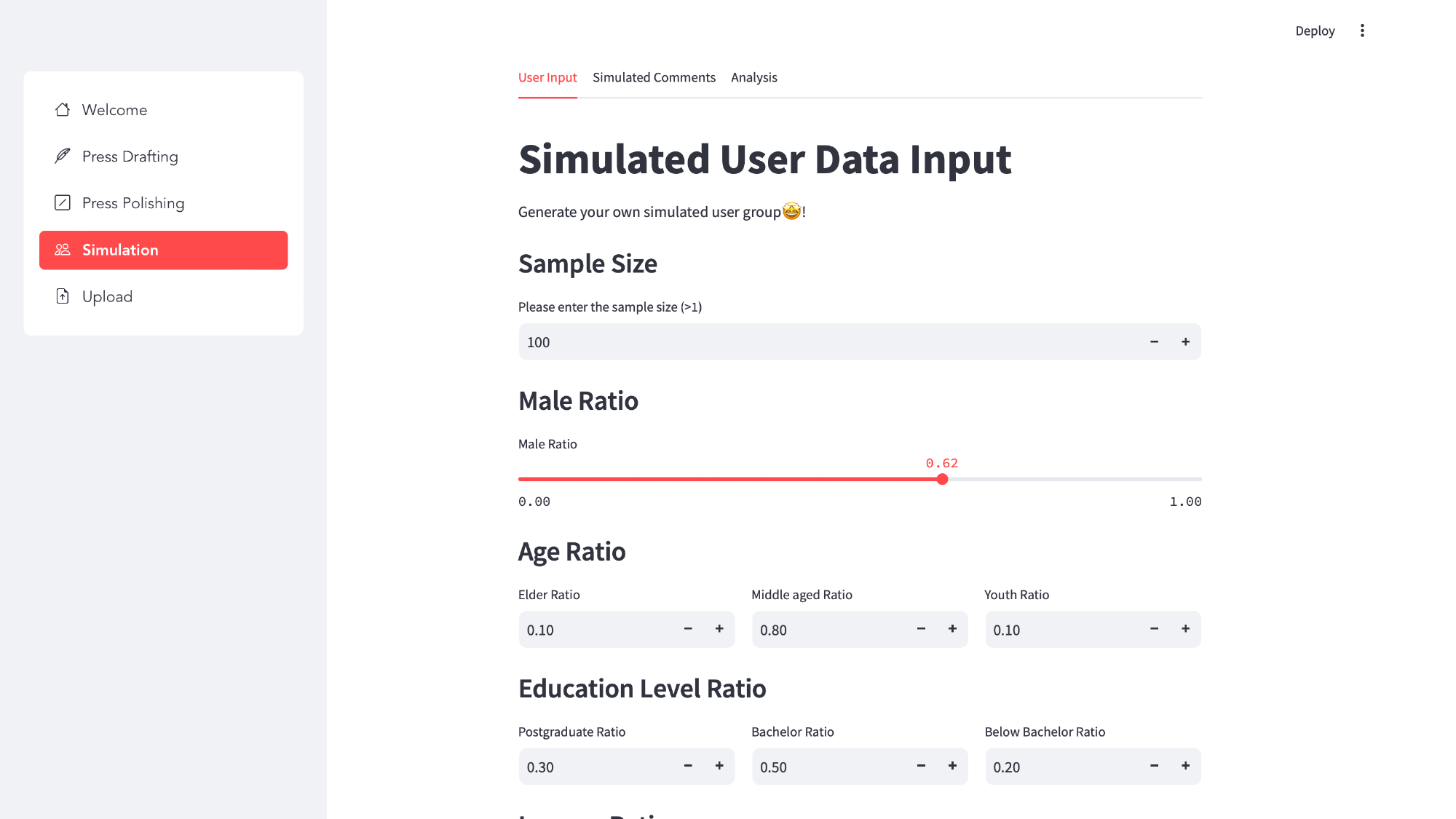}
\caption{A screenshot illustrates the user interface for generating a specific population distribution, utilizing customized demographic indicators.}
\label{fig:demo3-3}
\end{figure*}

\begin{figure*}[htbp]
\centering
\includegraphics[width=0.85\textwidth]{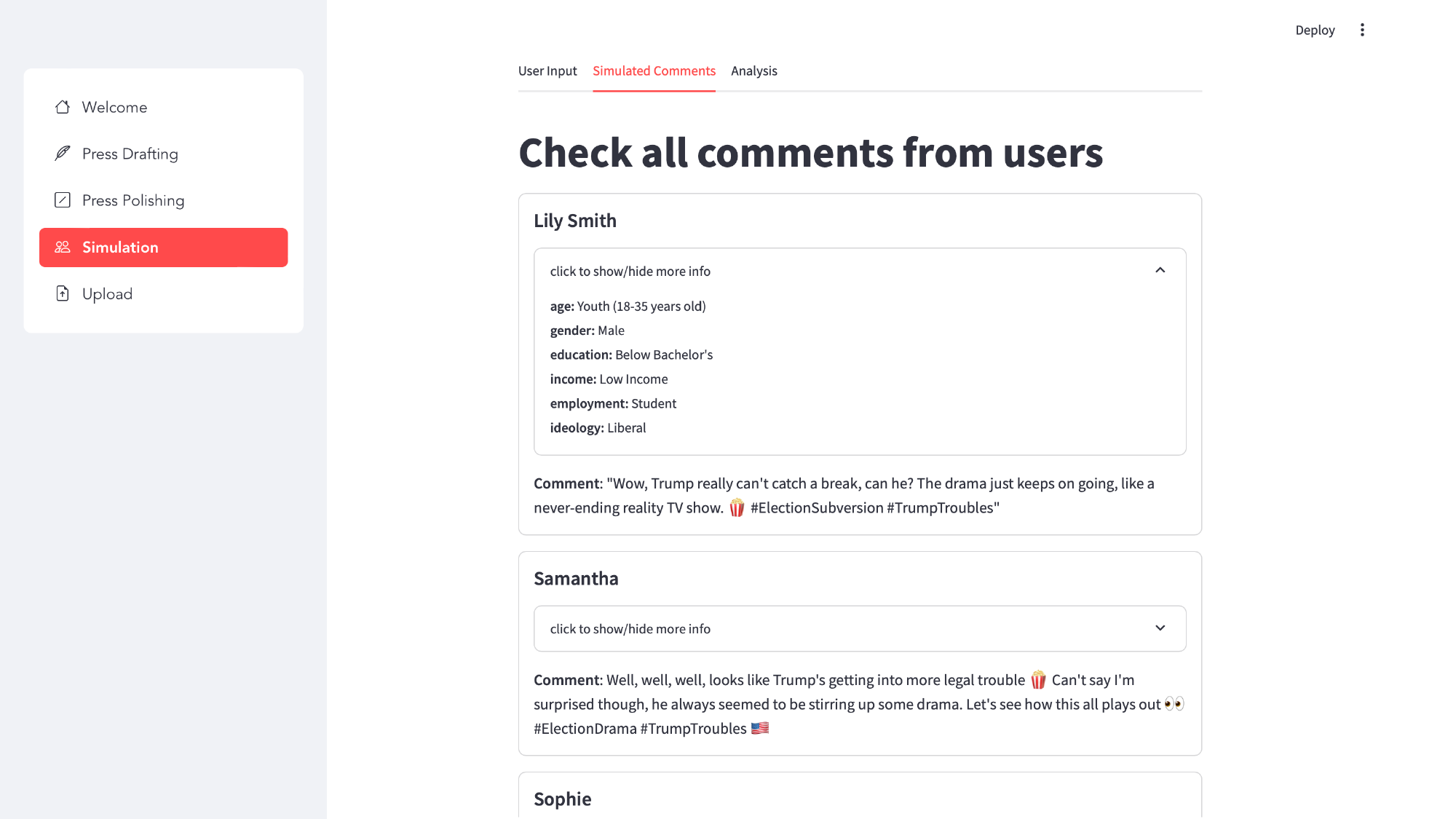}
\caption{A screenshot presents the generated simulated comments. The specific demographic indicators of the simulated commenters are also displayed.}
\label{fig:demo3-4}
\end{figure*}

\begin{figure*}[htbp]
\centering
\includegraphics[width=1\textwidth]{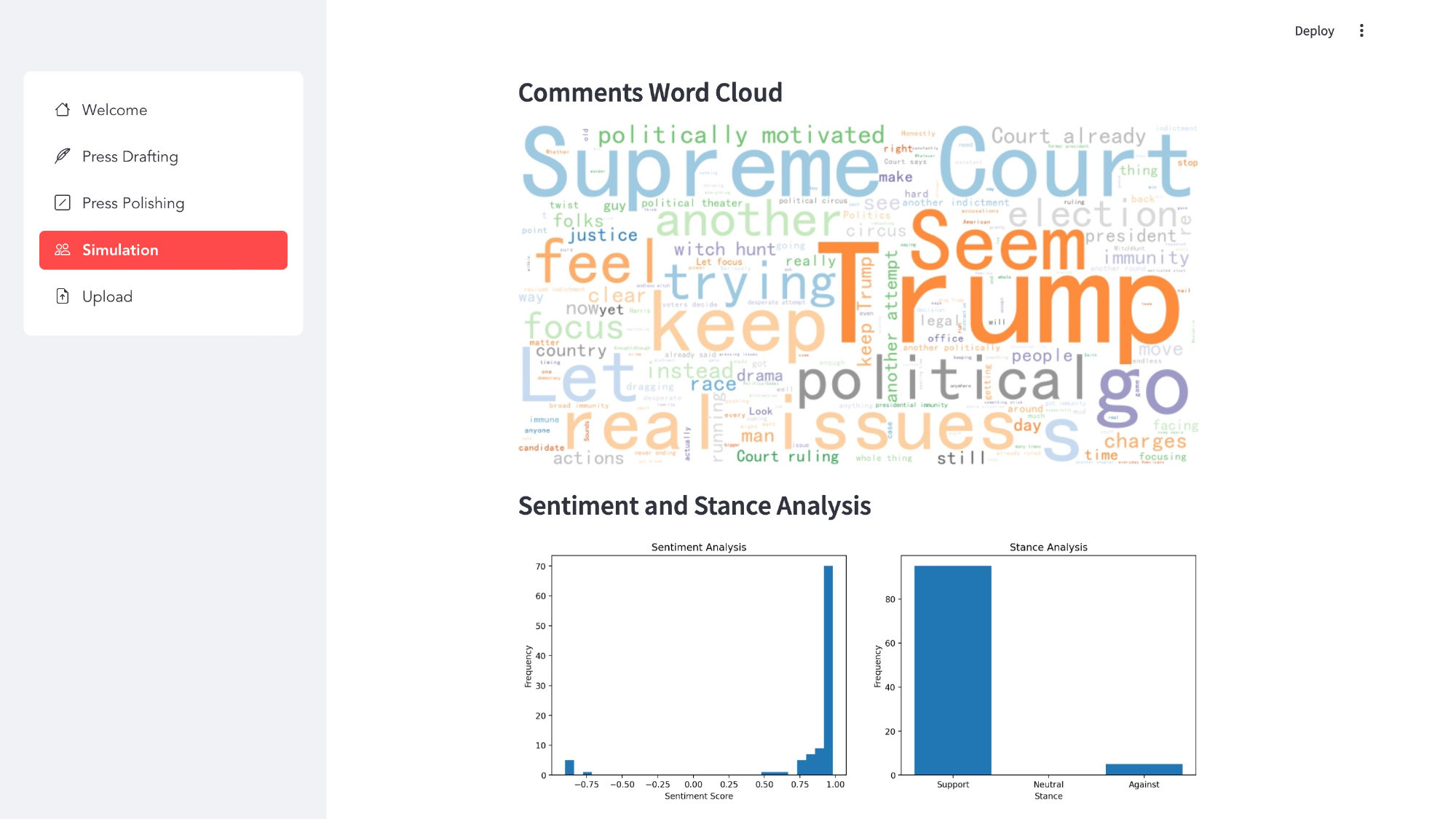}
\caption{The screenshot shows the analysis interface of simulated feedback, which displays the word cloud of comments and the frequency statistics of sentiments and stances.}
\label{fig:demo3-5}
\end{figure*}

\clearpage
\section{Prompts for LLMs}
\label{sec: Prompts for LLMs}

\begin{table*}[h!]
\centering
\begin{tabular}{@{}ll@{}}
\toprule
Genres     & Prompts                                                                                                                                                                                                                \\ \midrule
News       & \begin{tabular}[c]{@{}l@{}}Write a news release with a genre NEWS based on the following content.\\ News reports on current events with timeliness and objectivity.\\ content:\{content\}.\end{tabular}                \\ \midrule
Profile    & \begin{tabular}[c]{@{}l@{}}Write a news release with a genre of PROFILE based on the following content.\\ Profile journalism offers in-depth looks at people or topics.\\ content:\{content\}.\end{tabular}            \\ \midrule
Commentary & \begin{tabular}[c]{@{}l@{}}Write a news release with a genre of COMMENTARY based on the following content.\\ Commentary provides analysis and opinion on current events.\\ content:\{content\}.\end{tabular} \\ \bottomrule
\end{tabular}
\caption{Prompts for LLMs. \{content\} is news corpus.}
\label{tab:Prompts for LLMs}
\end{table*}

\section{Prompts for Agents on Press Drafting Module}

\subsection{Prompts for Searchers}
\label{sec:Prompts for Searchers}

\begin{table*}[h!]
\centering
\scalebox{0.88}{
\begin{tabular}{@{}ll@{}}
\toprule
Searchers & Prompts                                                                                                                         \\ \midrule
Searcher1 & \begin{tabular}[c]{@{}l@{}}\{content\}\\ Based on the above corpus, extract the core elements of the event, including time, place, key people, etc.\\ Here are the detailed instructions:\\ Identify the exact date and timeframe of the event.\\ Identify the exact location of the incident, including the city, region, or even a specific place.\\ Identify the key people involved, such as the dominant person, the victim, or the relevant authority.\end{tabular} \\     \midrule
Searcher2 & \begin{tabular}[c]{@{}l@{}}\{content\}\\ Sort out the passage of time timeline and key plot points based on the provided news corpus.\\ Describe the course of events and record important steps and twists in time sequential order.\\ Extract key episodes and details, e.g. flashpoints, important decisions, or actions.\\ For example, at YYYY-MM-DD, some events happened, etc.\end{tabular}                                                                                                        \\ \bottomrule
\end{tabular}
}
\caption{Prompts for Searchers. \{content\} is news corpus, the same as the ones in Table~\ref{tab:Prompts for LLMs}.}
\end{table*}

\clearpage
\subsection{Prompts for Writers}
\label{sec:Prompts for Writers}

\begin{table*}[h!]

\centering
\scalebox{0.83}{
\begin{tabular}{@{}cl@{}}
\toprule
Writers               & Prompts                                                                                       \\ \midrule
Title                 & \begin{tabular}[c]{@{}l@{}}\{content\}\\Based on the content of UserRequirement, extract the core elements, process, \\ key plot of the event, and the collected background information and impact.\\ Propose 3-5 headlines for the news.\\ Please return the result based on the following JSON structure: {[}\{\{"title": str\}\}{]}.\end{tabular}                                                           \\ \midrule
Content of News       & \begin{tabular}[c]{@{}l@{}}\{content\}\\Select the most suitable title from the Title.\\ A good news headline should be accurate, concise, and attractive.\\ Complete the writing of the press release and present a complete, professional, \\ excellent, and directly publishable press release.\\ Requirement:\\ - Refer to the language style, article structure, and narrative techniques of {[}News Database{]}, \\  maintain objectivity and neutrality, and present the occurrence, development, and results of events \\  in a clear structure. Use concise and clear language, avoiding lengthy and complex sentences \\  as well as obscure vocabulary.\\ - You can refer to the factual basis that may be used in the {[}Fact Database{]} (or not) to correct \\  the misinformation in the press release. If there are references, \\  please indicate the source at the end of the news.\\ - Based on the information from UserRequirement and the {[}Internet Surfer{]}\\  as the main basis and theme for your writing, use all facts as a benchmark and write \\  according to the format of news reports, including titles, introductions, main body, and endings.\end{tabular} \\ \midrule
Content of Profile    & \begin{tabular}[c]{@{}l@{}}\{content\}\\Select the most suitable title from the Title.\\ A good profile title for a profile should highlight its characteristics and charm.\\ Complete the writing of the character profile and present a complete, \\ vivid, and in-depth close-up of the character.\\ Requirement:\\ - Using information from UserRequirement and the {[}Internet Surfer{]} as the main materials,\\  the characters' personalities, achievements, and stories are presented through descriptions \\  of their appearance, personality, behavior, and language. By using detailed descriptions \\  and scene reproduction, readers can feel the true existence and emotional world of the characters.\\ - Referring to the language style, article structure, and narrative techniques of {[}News Database{]}.\\ - Referring to factual evidence from the {[}Fact Database{]} that can be used to enrich character images,\\  such as their experiences, achievements, and contributions. \\  Language expression should be delicate and emotional,\\  using appropriate adjectives and adverbs to enhance the character's infectiousness and affinity.\end{tabular}         \\ \midrule
Content of Commentary & \begin{tabular}[c]{@{}l@{}}\{content\}\\Select the most suitable title from the Title.\\ A good comment title should introduce the event directly with a viewpoint or appeal.\\ Complete the writing of news commentary and present a complete, professional, \\ and in-depth news commentary manuscript.\\ Requirement:\\ - Based on information from UserRequirement and the {[}Internet Surfer{]},\\  conduct an in-depth analysis of the background, causes, and impact of the event.\\ - Propose unique perspectives and analyses.\\  Use logical reasoning and evidence to support viewpoints, \\  avoiding subjective speculation and emotional expression.\\ - Referring to the language style, article structure, and narrative techniques of {[}News Database{]}, \\  it is appropriate to cite factual evidence from {[}Fact Database{]} to support viewpoints \\  and enhance the credibility and persuasiveness of comments.\\ - Language expression should be persuasive and infectious,\\  using vivid vocabulary and vivid metaphors to attract readers' attention and evoke resonance.\end{tabular}                                                                            \\ \bottomrule
\end{tabular}
}
\caption{Prompts for Writers. \{content\} contains user input, as well as the results of RAG: [News Database], [Fact Database], and [Internet Surfer].}
\end{table*}

\clearpage
\section{Prompts for Agents on Press Polishing Module}
\label{sec:Prompts for Agents on Press Polishing Module}

\subsection{Prompts for Reviewers}
\begin{table}[h!]
\centering
\scalebox{0.88}{
\begin{tabular}{@{}cl@{}}
\toprule
Genres     & Prompts                                                                                                                              \\ \midrule
News       & \begin{tabular}[c]{@{}l@{}}News: \{content\}\\ Please review the press release and provide critical comments based on these aspects:\\ - Timeliness: Whether the latest and most important information was reported on time?\\ - Accuracy: Whether the factual statements are accurate and reliable, \\   and whether the data and references are reliable?\\ - Concise and clear: Can the core information be conveyed clearly in concise language, \\   avoiding lengthy and complex expressions?\\ - Key emphasis: Have the key elements and focus of the event been clearly identified?\end{tabular}                                                                                                                                                            \\ \midrule
Profile    & \begin{tabular}[c]{@{}l@{}}Profile: \{content\}\\ Please review the profile and provide critical comments based on these aspects:\\ - Unique perspective: Whether a unique and innovative angle has been chosen to \\   present the theme or characters?\\ - Detail description: Whether it contains vivid and specific details that enable readers \\   to have a strong sensory experience?\\ - Personalization: Can the personality traits of the theme or character be highlighted \\   to distinguish it from other similar individuals?\end{tabular}                                                                                                                                                                                                          \\ \midrule
Commentary & \begin{tabular}[c]{@{}l@{}}Commentary: \{content\}\\ Please review the commentary news article and provide critical comments based on these aspects:\\ - Depth and breadth: Whether the topic has been analyzed comprehensively and in-depth, \\   with extensive background information and details presented?\\ - Sufficiency of evidence: Whether sufficient facts, data, cases, \\   and other evidence are provided to support the viewpoints and arguments?\\ - Clarity of opinions: Whether viewpoints are expressed clearly and distinctly, \\   avoiding ambiguous language and ensuring that the stance taken is easily understandable?\\   Whether the main argument is stated upfront and supported by coherent explanations and examples?\end{tabular} \\ \bottomrule
\end{tabular}
}
\caption{Prompts for Reviewers. \{contents\} is the input news draft.}
\end{table}

\subsection{Prompts for Rewriters}
\begin{table}[h!]
\centering
\begin{tabular}{@{}l@{}}
\toprule
\begin{tabular}[c]{@{}l@{}}News Draft: \{news draft\}\\ Comments: \{comments\}\\ You have received the following comments on the news draft. Please revise the news accordingly and \\ provide a revised version of the news.\\ Just return revised news. Do not explain your reasoning.\end{tabular} \\ \bottomrule
\end{tabular}
\caption{Prompts for Rewriters.}
\end{table}

\clearpage
\section{Prompts for Simulation Module}
\label{sec:Prompts for Simulation Module}

\begin{table}[h!]
\centering
\scalebox{0.89}{
\begin{tabular}{@{}l@{}}
\toprule
\begin{tabular}[c]{@{}l@{}}You are a \{gender\} who is \{age\}, with your highest education \{education\}. Your income level is \{income\},\\ and your current employment status is \{employment\}. You tend to be \{ideology\} when making comments.\\ \\ Today you saw a news article as follows:\\ \\ news article: \{news\}\\ You want to post your own comments in the comment section below the news article.\\ Your comment doesn't have to be formal. It can be as casual as you want—use slang, emojis, or even a bit of sarcasm.\\ Please make sure your comment conveys a perspective consistent with your role configuration.\\ \\ Here is what you have posted in the past: \{historical\_comment\}\\ \\ You can refer to your past tone and wording habits when posting your comment,\\ but do not mention the events in your historical comment unless it is highly relevant to the current news.\\ \\ Reply with your authentic voice:\end{tabular} \\ \bottomrule
\end{tabular}
}
\caption{Prompts for comments simulation.}
\end{table}

\clearpage
\section{User Profile Pool Generating}
\label{sec:Prompts for User Profile Pool Generating}

\subsection{User Data}
The user data we utilize originates from the Twitter social platform and is completely anonymized. We categorize users based on their publicly available historical tweets, ensuring compliance with all privacy regulations.

\subsection{Prompts for User Annotation}
\label{sec:Prompts for User Annotation}

\begin{table}[h!]
\centering
\scalebox{0.9}{
\begin{tabular}{@{}ll@{}}
\toprule
\begin{tabular}[c]{@{}l@{}}You are a user content analyst tasked with determining various attributes of a user based on the content they post.\\ Your goal is to categorize these attributes into several distinct categories.\\ Please strictly follow the options provided for selection, and do not return null.\\ \\ \textbf{Demographic Attributes}:\\ - Age: Inferred from mentions of life stages, time markers, references to popular culture, etc.\\ 1. Youth (18-35 years old)\\ 2. Middle-aged (36-65 years old)\\ 3. Elderly (over 65 years old)\\ \\ - Gender: Inferred from self-references, pronouns, interests, lifestyle details, etc.\\ 1. Male\\ 2. Female\\ \\ - Income: Inferred from mentions of living standards, spending habits, occupation, etc.\\ 1. Low income\\ 2. Middle income\\ 3. High income\\ \\ - Education: Inferred from vocabulary, sentence structure, mentions of educational background, etc.\\ 1. Below Bachelor's\\ 2. Bachelor's degree\\ 3. Postgraduate education\\ \\ - Employment: Inferred from descriptions of daily activities, work environment, professional terminology, etc.\\ 1. Working now\\ 2. Student\\ 3. Others\\ \\ - Ideology: Inferred from political views, party support, election tendencies, etc.\\ 1. Liberal\\ 2. Moderate\\ 3. Conservative\\ \\ Output in the following JSON format:\\ \{\{"age": str, "gender": str, "Income": str, "Education": str, "employment": str, "Ideology": str\}.\\ \\ \textbf{User Content}:\\ user name: \{name\}\\ user posts:\{content\_list\}\\ \\ \textbf{Output}:\end{tabular} &  \\ \bottomrule
\end{tabular}
}
\caption{Prompts for user annotation.}
\end{table}

\clearpage
\section{Evaluation Experiment}
\label{sec:Evaluation Experiment}

\subsection{Prompts for GPT4o Scoring on Press Generating Experiment}
\label{sec:Prompts for GPT4o Scoring on Press Generating Experiment}
\begin{table}[h!]
\centering
\scalebox{0.75}{
\begin{tabular}{@{}ll@{}}
\toprule
Genres     & Prompts                                                                                                                                                                                             \\ \midrule
News       & \begin{tabular}[c]{@{}l@{}}\{content\}\\- For comprehensiveness: Consider whether the news provides a comprehensive overview of the event, \\ including various aspects such as background, development process, key figures involved, and possible consequences. \\ Also, assess if it presents different perspectives and viewpoints related to the event \\ to give readers a more complete understanding.\\ - For depth: Evaluate whether the news goes beyond the surface of the event and delves into deeper issues \\ such as underlying causes, long-term impacts, and potential solutions. \\ Also, check if it provides in-depth analysis and insights through interviews with experts or research on related topics.\\ - For objectivity: Determine if the news report is free from personal biases and subjective views and presents all aspects\\ of the event in a neutral manner. Also, see if it gives equal expression opportunities to different \\ viewpoints and stakeholders without favoring any one side.\\ - For importance: Consider whether the event covered by the news has significant social, political, economic, \\ or cultural significance. Also, assess if the news delves into the reasons and impacts \\ behind the event to provide valuable analysis and thinking.\\ - For readability: Evaluate whether the language of the news is clear, concise, and easy to understand, avoiding overly\\ technical or rare vocabulary and complex sentence structures. \\ Also, check if the structure of the news is reasonable with a clear lead, body, and conclusion and if the logic is coherent.\end{tabular}                                                                                                                                                                       \\ \midrule
Profile    & \begin{tabular}[c]{@{}l@{}}\{content\}\\- For richness: The profile contains abundant details that vividly depict the person's appearance, \\ habits, and work environment, making the character more three-dimensional. \\ It presents the person from multiple perspectives, including evaluations from family, friends, colleagues, and partners, \\ providing a comprehensive understanding. The story has a clear beginning, development, and end, with a coherent plot. \\ It also includes a variety of elements such as challenges faced, solutions found, and achievements attained to enrich the narrative.\\ - For depth: Thoroughly analyze the person's inner world, motives, and values. \\ Show the person's growth and changes at different stages. \\ Uncover the unknown stories and experiences behind the person to enrich the character's image.\\ - For uniqueness: Approach the person's story from a unique perspective, different from common ways of reporting on people.\\  Highlight the person's individuality and distinctive features. Be able to discover and show the little-known side of the person, \\ bringing freshness to readers.\\ - For inspiration: The person's story can inspire emotional resonance in readers, such as admiration, touch, and inspiration. \\ Let readers gain inspiration and positive energy from the person. Through vivid descriptions and narratives, \\ readers can deeply feel the charm and influence of the person.\\ - For readability: The language expression is smooth, vivid, and easy to understand. \\ Use appropriate description methods and narrative techniques to enhance the attractiveness of the article. \\ Have a reasonable structure, clear hierarchy, and logical clarity for easy reading and understanding.\end{tabular} \\ \midrule
Commentary & \begin{tabular}[c]{@{}l@{}}\{content\}\\- For comprehensiveness: Cover all aspects of the topic thoroughly, including different viewpoints, \\ potential consequences, and historical background. \\ Incorporate a wide range of sources and examples to provide a holistic understanding.\\ - For clarity of opinions: Express viewpoints clearly and precisely, avoiding ambiguity and vagueness. \\ Use straightforward language and well-structured arguments to make the stance easily discernible.\\ - For sufficiency of evidence: The news has sufficient facts, data, cases, and other evidence to support a viewpoint, \\ making it more persuasive.\\ - For relevance: Be closely related to current events and issues of significance. \\ Address topics that are of interest and concern to the audience.\\ - For readability: The language expression is smooth, vivid, and easy to understand. \\ Use appropriate description methods and narrative techniques to enhance the attractiveness of the article. \\ Have a reasonable structure, clear hierarchy, and logical clarity for easy reading and understanding.\end{tabular}                                                                                                                                                                                                \\ \bottomrule
\end{tabular}
}
\caption{Prompts for GPT4o Scoring on Press Generating Experiment. \{content\} is the press that needs to be scored.}
\end{table}

\clearpage
\subsection{Experiment Press Release Introduction}
\label{sec:Experiment Press Release Introduction}
\begin{table}[h!]
\centering
\scalebox{1}{
\begin{tabular}{@{}lll@{}}
\toprule
Genre      & number & fields                                                                                                                           \\ \midrule
News       & 100    & \begin{tabular}[c]{@{}l@{}}International, technology, art, sports, health, tourism, real estate, fashion, etc\end{tabular} \\ \midrule
Profile    & 100    & \begin{tabular}[c]{@{}l@{}}International, domestic, book, sports, film, etc\end{tabular}                                      \\ \midrule
Commentary & 100    & \begin{tabular}[c]{@{}l@{}}International, art, sports, business, technology, etc\end{tabular}                                 \\ \bottomrule
\end{tabular}
}
\caption{Experiment Press Release Introduction}
\label{tab:data}
\end{table}

\subsection{Simulation Experiment Press Release Introduction}
\label{sec:Simualtion Experiment Press Release Introduction}
\begin{table}[h!]
\centering
\scalebox{1}{
\begin{tabular}{@{}lll@{}}
\toprule
Field      & No. of comments & Description                                                                                                                           \\ \midrule
Politics       & 525    & \begin{tabular}[c]{@{}l@{}}Questions regarding Trump's age and capacity\end{tabular} \\ \midrule
Economy    & 253    & \begin{tabular}[c]{@{}l@{}}The challenges of offshore wind\end{tabular}                                      \\ \midrule
Conflict & 131    & \begin{tabular}[c]{@{}l@{}}Conflicts between Israel and Hezbollah\end{tabular}                                 \\ \bottomrule
\end{tabular}
}
\caption{Simulation Experiment Press Release Introduction}
\label{tab:data}
\end{table}

\subsection{Prompts for GPT4o on Simulation Experiment}
\label{sec:Prompts for GPT4o on Simulation Experiment}
\begin{table}[h!]
\centering
\scalebox{0.89}{
\begin{tabular}{@{}l@{}}
\toprule
\begin{tabular}[c]{@{}l@{}}You are a user review analyst responsible for judging the emotional tendency of user comments, \\ and determining the user's viewpoint and stance based on the provided news articles and user comments.\\ The viewpoint of the news article is as follows: \{news\}\\ The comments posted by users under the news article are as follows:\{comment\}\\ \\ \textbf{Sentiment inclination}: \\ Please determine whether the sentiments expressed in the comments posted by the user are positive, negative, or neutral.\\ - Positive\\ - Neutral\\ - Negative\\ \\ \textbf{Sentiment score}:\\ Please provide a score range of {[}-1,1{]} based on the comments posted by the user.\\ The closer the score is to -1, the more negative it is, and the closer the score is to 1, the more positive it is.\\ A score of 0 indicates neutrality.\\ The given score is between -1 and 1, with two decimal places retained.\\ \\ \textbf{Stance}: \\Please judge whether the user supports, opposes, or is neutral towards the news viewpoint\\ based on the comments posted by the user and the viewpoint of the news article.\\ - Support\\ - Neutral\\ - Against\\ \\ Output in the following JSON format:\\ \{\{"Sentiment\_inclination": "Positive", "Sentiment\_score":0.98, "Stance": "Support"\}\}\\ Please only return the JSON string.\\ Please strictly follow the options provided for selection, and do not return null.\end{tabular} \\ \bottomrule
\end{tabular}
}
\caption{Prompts for GPT4o on Simulation Experiment.}
\end{table}

\clearpage
\section{Case Study}
\label{sec:Case Study}

\subsection{Input Texts}
\begin{tcolorbox}[width=\linewidth]
Published Sept.9,2024 Updated Nov. 6, 2024

Leer en espanol

The last time the nation held a debate with the presidency on the line, a candidate with about eight decades of life behind him faced the challenge of proving that he was still up to the job of running the country. He failed.

Two and a half months later, the cast of characters has shifted and another candidate heading toward the octogenarian club confronts his own test to demonstrate that he has not diminished with ageWhether he passes that test may influence who will be the next occupant of the Oval Office.
\end{tcolorbox}

\subsection{Press Drafting}
News Release

*Published: Sept. 9, 2024 | Updated: Nov. 6, 2024*

Title: Age and Leadership: The Growing Challenge in the Presidential Race

As the 2024 U.S. Presidential Election approaches, the spotlight is once again on the age of the candidates vying for the highest office in the land. The last presidential debate, held in late 2020, saw a candidate in his late 70s struggling to prove his vitality and readiness to lead the country. Despite his decades of political experience, he ultimately fell short in convincing voters that he could meet the demands of the presidency, a role that requires both mental acuity and physical stamina.

Fast forward to 2024, and the situation has shifted. As the country gears up for another election, a new contender, also approaching his 80th birthday, now faces the same daunting challenge. Like his predecessor, this candidate is tasked with demonstrating that his advanced years do not hinder his ability to lead. His performance in the upcoming debates and campaign events will be closely scrutinized, as it could play a pivotal role in determining the next occupant of the Oval Office.

Voter concerns over age-related issues such as mental sharpness, physical endurance, and the ability to adapt to the rapidly changing world are expected to be central themes in the race. Both political parties are carefully navigating these concerns, as the outcome could significantly influence the trajectory of the election.

As the debates approach, analysts and political strategists are keenly aware of the growing impact of age on the national conversation. Whether this new candidate can meet the test of leadership in the eyes of the American people could ultimately decide the future direction of the country.

For further information, please contact:
[Your Media Contact Information]

---

*Leer en Español*

\subsection{Press Polishing}
News Release

*Published: Sept. 9, 2024 | Updated: Nov. 6, 2024*

Title: Age and Leadership: The Challenge Facing 2024 Presidential Candidates

As the 2024 U.S. Presidential Election heats up, age once again looms as a key factor in the race. In 2020, a candidate nearing his 80s struggled to prove his mental and physical readiness for the demands of the presidency, and despite decades of experience, he was unable to convince voters that he could lead the nation effectively.

Now, in 2024, another candidate approaching his 80th birthday is set to face the same scrutiny. Like his predecessor, he must demonstrate that age does not diminish his ability to lead the country. With the election just around the corner, his performance in upcoming debates and campaign events could play a crucial role in shaping the next administration.

Voter concerns about mental sharpness, physical stamina, and adaptability in a rapidly evolving world are expected to be central themes in the campaign. Both political parties are acutely aware of the importance of addressing these issues, as they could significantly influence the outcome of the election.

As debates and events unfold, political analysts are closely monitoring how the candidates will handle these age-related challenges. Whether this candidate can prove his leadership abilities in the eyes of the American people may very well determine the future direction of the country.

For further information, please contact:
[Your Media Contact Information]

---

*Leer en Español*

---

Changes Made:

1. Timeliness: Removed the reference to "the last time the nation held a debate with the presidency on the line" and replaced it with a more direct reference to 2020, improving timeliness and relevance.

2. Conciseness: Streamlined sentences for better clarity and flow, especially regarding the previous candidate’s failure to convince voters and the challenges facing the new candidate.

3. Key emphasis: Strengthened the focus on age as a central concern for both candidates, emphasizing its potential impact on the election outcome.

4. Clarity: Simplified complex phrases such as "whether he passes that test may influence who will be the next occupant of the Oval Office" to "his performance could play a crucial role in shaping the next administration."

\subsection{Simulation}
John Harris
\begin{tcolorbox}[width=\linewidth]
"Age is a factor in any race, but the real test is whether these candidates can handle the mental and physical demands of the presidency. This election will be a true test of their leadership abilities."
\end{tcolorbox}

Sarah Jenkins, Voter from Ohio  
\begin{tcolorbox}[width=\linewidth]
"I respect their careers, but I'm concerned about whether older candidates can meet the demands of the job. It’s about being realistic, not disrespectful."
\end{tcolorbox}

Carlos Rivera
\begin{tcolorbox}[width=\linewidth]
"Experience is valuable, but how they perform in debates and public events will be crucial. If they prove their vitality, age won’t be an issue."
\end{tcolorbox}

Emily Thompson
\begin{tcolorbox}[width=\linewidth]
"As a young person, I worry that older candidates are disconnected from issues that matter to my generation. We need a leader who gets it."
\end{tcolorbox}

David Mitchell
\begin{tcolorbox}[width=\linewidth]
"Age shouldn’t disqualify anyone. Leadership is about character and decisions, not just how old you are."
\end{tcolorbox}

\clearpage


\end{document}